\documentclass[letterpaper]{article}
\usepackage{aaai}
\usepackage{times}
\usepackage{helvet}
\usepackage{courier}
\usepackage{array}
\usepackage{booktabs}
\usepackage{multirow, booktabs, array}
\usepackage{graphicx}
\usepackage{subfigure}
\usepackage{parskip}
\usepackage{footmisc}
\usepackage{amsmath}
\usepackage{amssymb}
\usepackage[square]{natbib}
\usepackage{hyperref}
\usepackage{url}
\makeatletter
\def\NAT@aysep{} 
\makeatother

\nocopyright 

\begin{document}

\title{VLM-3D: End-to-End Vision-Language Models for Open-World 3D Perception}
\author{
    Fuhao Chang\textsuperscript{1,2,3,$\ast$}
    Shuxin Li\textsuperscript{1,2,4,}\thanks{These authors contributed equally to this work},
    Yabei Li\textsuperscript{1,2,5},
    Lei He\textsuperscript{1,2}\thanks{Corresponding author: helei2023@tsinghua.edu.cn} \\
    \textsuperscript{1}School of Vehicle and Mobility, Tsinghua University, Beijing, 100084, China \\
    \textsuperscript{2}State Key Laboratory of Intelligent Green Vehicle and Mobility, Tsinghua University, Beijing, 100084, China \\
    \textsuperscript{3}College of Information and Electrical Engineering, China Agricultural University, Beijing, 100091, China \\
    \textsuperscript{4}School of Statistics and Data Science, Southwestern University of Finance and Economics, Chengdu, 611130, China \\
    \textsuperscript{5}Meituan, China
}
\maketitle

\begin{abstract}
Open-set perception in complex traffic environments poses a critical challenge for autonomous driving systems, particularly in identifying previously unseen object categories, which is vital for ensuring safety. Visual Language Models (VLMs), with their rich world knowledge and strong semantic reasoning capabilities, offer new possibilities for addressing this task. However, existing approaches typically leverage VLMs to extract visual features and couple them with traditional object detectors, resulting in multi-stage error propagation that hinders perception accuracy. To overcome this limitation, we propose VLM-3D, the first end-to-end framework that enables VLMs to perform 3D geometric perception in autonomous driving scenarios. VLM-3D incorporates Low-Rank Adaptation (LoRA) to efficiently adapt VLMs to driving tasks with minimal computational overhead, and introduces a joint semantic-geometric loss design: token-level semantic loss is applied during early training to ensure stable convergence, while 3D IoU loss is introduced in later stages to refine the accuracy of 3D bounding box predictions. Evaluations on the nuScenes dataset demonstrate that the proposed joint semantic-geometric loss in VLM-3D leads to a 12.8\% improvement in perception accuracy, fully validating the effectiveness and advancement of our method.
\end{abstract}

\section{Introduction}
Autonomous driving systems demand robust perception capabilities to safely navigate complex and dynamic real-world traffic environments. Among these capabilities, open-set perception\citep{huang2023measuring} the ability to recognize and detect previously unseen object categories is particularly critical for ensuring driving safety, especially in long-tail scenarios and anomalous traffic conditions. Conventional methods, which rely heavily on predefined category labels \citep[]{liu2024ov3d}, struggle to generalize to unseen classes, resulting in frequent missed detections or false alarms in dynamic settings, thereby significantly increasing the risk of traffic accidents. Consequently, developing open-set perception systems capable of real-time identification and classification of both known and unknown objects is paramount for advancing the safety and intelligence of autonomous driving technologies \citep[]{li2025u2ad}.

Visual Language Models (VLMs), with their powerful cross-modal representation abilities and rich semantic knowledge \citep[]{tong2024eyes}, have emerged as promising tools for open-set object detection. Trained on large-scale vision-language datasets, VLMs effectively align visual and textual features, enabling zero-shot detection capabilities \citep[]{zhao2024vlm}. However, existing approaches typically utilize VLMs as fixed feature extractors combined with traditional 2D or 3D detection models in multi-stage pipelines \citep[]{zhu2023pointclip}. This architectural separation leads to error propagation across stages, limiting joint optimization potential and constraining fine-grained 3D spatial reasoning—an essential aspect of autonomous driving perception stacks \citep{li2025u2ad}. Moreover, the lack of end-to-end integration further impedes their practical deployment in real-world scenarios \citep[]{yin2024survey}.

To address these limitations, we propose VLM-3D, an end-to-end 3D open-set perception framework built upon the Qwen2-VL visual language model. By fusing autonomous driving images and textual information, VLM-3D delivers an efficient and robust 3D geometric perception system \citep[]{zang2025contextual}. We incorporate Low-Rank Adaptation (LoRA) within the visual language backbone to enable highly efficient fine-tuning with minimal parameter overhead, ensuring a lightweight model suitable for deployment on embedded in-vehicle platforms \citep[]{xu2025lowrank, hu2022lora}. Additionally, we design a novel joint semantic-geometric loss strategy: token-level semantic loss is applied during early training stages to ensure stable convergence, while 3D Intersection-over-Union (IoU) loss is progressively introduced in later stages to refine the accuracy of 3D bounding box predictions \citep[]{yin2021centerpoint}.

Our main contributions include:
\begin{itemize}
\item Employing LoRA to efficiently fine-tune the Qwen2-VL model by integrating it into self-attention modules, achieving effective fusion of image and text modalities and directly predicting 3D bounding boxes in the LiDAR coordinate system. This approach significantly reduces computational costs, enabling deployment on embedded autonomous driving devices while maintaining high detection accuracy.

\item Proposing a novel two-stage loss function design that balances semantic alignment with geometric precision. The first stage uses mean squared error (MSE) loss to align multimodal features with semantic labels for rapid and stable convergence. The second stage introduces 3D IoU loss to precisely optimize bounding box center, size, and orientation, substantially enhancing geometric prediction accuracy. Dynamic weighting of these losses ensures optimal perception performance across diverse traffic scenarios.

\item Developing an end-to-end VLM-3D system architecture that unifies data preprocessing, modality fusion, and training optimization within a single framework. Unlike existing multi-stage pipelines, this design minimizes information loss, enhances system robustness, and improves deployment efficiency, demonstrating superior practicality and scalability for complex, real-world autonomous driving environments.

\end{itemize}

\section{Ralated work}
Traditional closed-set detection methods for autonomous driving leverage deep learning models trained on sensor data, such as point clouds and images, for object recognition and localization. These methods assume that object categories in the testing phase are known during training. Notable frameworks include VoxelNet, proposed by \citet{zhou2018voxelnet} \citep{zhou2018voxelnet}, which voxelizes point cloud data and uses 3D convolutional networks for end-to-end detection, significantly improving accuracy. PointPillars, introduced by \citet{lang2019pointpillars} \citep{lang2019pointpillars}, encodes point clouds into pseudo-images and employs 2D convolutional networks to enhance computational efficiency, meeting real-time requirements. PointRCNN, developed by \citet{shi2019pointrcnn} \citep{shi2019pointrcnn}, uses a two-stage process to generate and refine 3D object proposals from point clouds, improving small object detection in complex scenes. CenterPoint, proposed by \citet{yin2021centerpoint} \citep{yin2021centerpoint}, simplifies detection by predicting 3D bounding boxes based on object centroids, enhancing multi-object tracking. While effective in specific scenarios like vehicle and pedestrian detection, these methods struggle with open-world complexity, lacking generalization to unknown categories and leading to missed or incorrect detections.

Open-set object detection aims to enable models to accurately recognize known categories, detect and classify unknown objects, and dynamically learn new categories without extensive retraining. Existing algorithms include OpenFMNav, which uses vision-language models to build semantic maps for zero-shot indoor navigation but suffers from high latency due to multi-module processing \citep[]{wang2024openfmnav}. OV3D, proposed by \citet{liu2024ov3d} \citep{liu2024ov3d}, employs images as a bridge for point cloud-text alignment, supporting open-vocabulary 3D semantic segmentation, but its indoor focus and non-end-to-end pipeline limit its applicability. VL-SAM, introduced by \citet{zhang2024vlsam} \citep{zhang2024vlsam}, leverages attention maps for training-free open-set detection, but its slow inference and susceptibility to VLM hallucinations restrict its use. CLIP-based open-set detection relies on image-text contrastive learning for zero-shot detection but is constrained by high annotation costs \citep[]{zhang2022clip}. PointCLIP extends CLIP to point clouds for 3D open-set detection, but its multi-stage processing limits its applicability in autonomous driving \citep[]{zhu2023pointclip}. These methods, primarily focused on indoor scenarios, rely on non-end-to-end pipelines, incur high computational costs, and struggle to meet the real-time demands of automatic driving.

Multimodal large models offer new opportunities for open-set perception by integrating visual, linguistic, and other modalities for robust semantic understanding. Qwen2-VL, an open-source vision-language model, supports joint processing of images, videos, and text, with real-time interaction and multilingual capabilities. CLIP builds a unified embedding space through large-scale image-text contrastive learning but struggles with direct 3D point cloud processing \citep[]{zhang2022clip}. LLaVA, with vision and language instruction tuning, enhances generalization but requires significant computational resources, limiting its use on embedded devices \citep[]{chen2024language}. These models face challenges in open-set perception, including high resource consumption, discrete outputs unsuited for continuous 3D spaces, and reliance on large-scale paired data.

To address these challenges, this study proposes an end-to-end 3D open-set perception framework based on Qwen2-VL, leveraging low-rank adaptation (LoRA) for fine-tuning and integrating point cloud, image, and text data for end-to-end 3D bounding box generation. The framework employs multimodal feature fusion and two-stage loss optimization to enhance real-time performance and robustness, with innovations including end-to-end design, dynamic feature fusion, weakly-supervised learning, incremental learning, and enhanced unknown object detection.

\section{Method}
\begin{figure*}[t]
\centering
\includegraphics[width=0.9\textwidth]{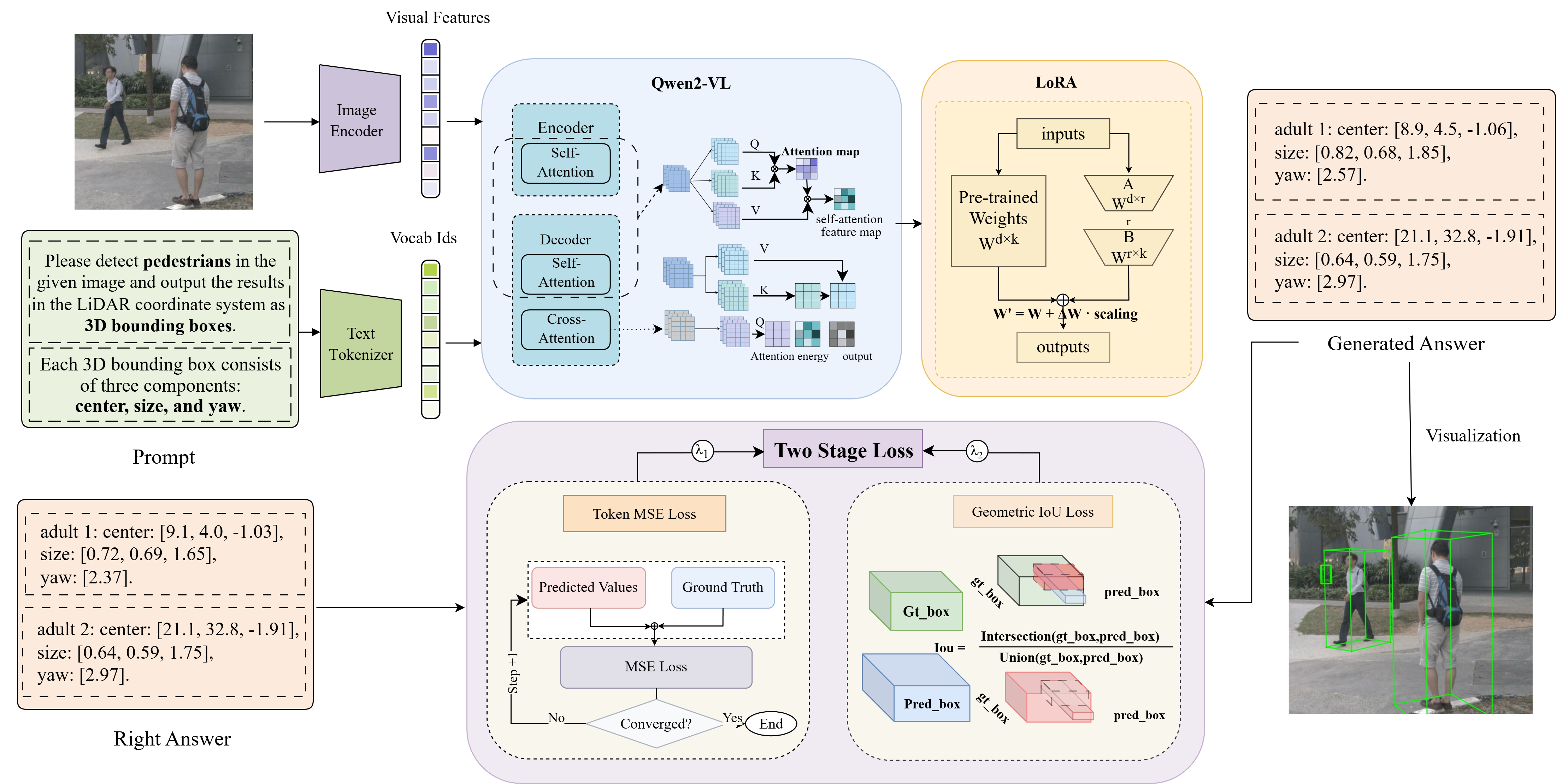}
\caption{The technical pipeline figure illustrates the 3D open-set perception framework based on Qwen2-VL, encompassing multimodal input preprocessing, LoRA-based feature fusion, two-stage loss optimization, and open-set strategy modules.}
\label{fig:pipeline}
\end{figure*}

This section presents a detailed description of the proposed end-to-end 3D open-set perception framework for complex traffic scenarios, addressing challenges in cross-modal semantic alignment and computational efficiency. The framework integrates the Qwen2-VL vision-language model and comprises four core components: multimodal input preprocessing, feature fusion with low-rank adaptation (LoRA), two-stage loss optimization, and an open-set strategy. These components collectively ensure efficient processing, robust semantic alignment, and high-precision 3D bounding box generation in the LiDAR coordinate system, tailored for real-time autonomous driving applications.

\subsection{Multimodal Input Preprocessing}
The preprocessing stage integrates and structures multimodal inputs to establish a robust foundation for subsequent feature fusion and inference. It consists of three sub-components, each designed to optimize data handling and ensure compatibility with the Qwen2-VL model.

\subsection{Input Data Reception and Preprocessing}

The framework receives two primary inputs: a user-provided text prompt and a task-related image. The text prompt, e.g., ``Detect pedestrians in the provided image and output results as 3D bounding boxes in the LiDAR coordinate system," is processed using natural language processing (NLP) techniques. Specifically, the prompt is tokenized using a pretrained tokenizer (e.g., based on the BERT architecture) and converted into vocabulary identifiers (Vocab IDs), transforming the text into a structured numerical representation suitable for semantic processing. The tokenizer employs subword tokenization to handle out-of-vocabulary terms, ensuring robust parsing of diverse prompts. The resulting Vocab IDs are embedded into a high-dimensional space using the Qwen2-VL text encoder, producing text features \( F_t \in \mathbb{R}^{d_t} \), where \( d_t \) is the text embedding dimension (typically 768 for Qwen2-VL).This structured numerical representation enables efficient semantic processing for subsequent tasks. The embedded text features are subsequently integrated with image features to facilitate accurate object detection and localization.

The input image, typically a 2D RGB frame from a vehicle-mounted camera, is processed using a convolutional neural network (CNN) backbone (e.g., ResNet-50 or Vision Transformer) integrated within Qwen2-VL. The image is resized to a fixed resolution (e.g., 224x224 pixels) and normalized to ensure consistency across inputs. The CNN extracts visual features \( F_v \in \mathbb{R}^{d_v} \), where \( d_v \) is the visual embedding dimension, capturing spatial structures, textures, and semantic cues relevant to pedestrian detection. To enhance efficiency, image preprocessing includes data augmentation techniques (e.g., random cropping, flipping, and color jittering) during training to improve model robustness. This step ensures that both text and visual inputs are standardized, enabling efficient integration into the Qwen2-VL model for subsequent multimodal processing.

\subsection{Task Definition and Output Format}
The task is defined as detecting pedestrians in the input image and generating 3D bounding boxes in the LiDAR coordinate system, aligning with the requirements of autonomous driving and robotic navigation. Each 3D bounding box is parameterized by:
- Center point (\( x, y, z \)): The 3D coordinates representing the pedestrian's position in the LiDAR coordinate system, measured in meters.
- Size (\( l, w, h \)): The length, width, and height of the bounding box, reflecting the pedestrian's physical dimensions.
- Yaw angle (\( \theta \)): The orientation angle in radians, indicating the pedestrian's heading relative to the LiDAR coordinate system.

The output format is standardized as a 7-dimensional vector \( [x, y, z, l, w, h, \theta] \) per detected pedestrian, ensuring compatibility with downstream tasks such as path planning and collision avoidance. This parameterization aligns with common 3D object detection datasets (e.g., KITTI, nuScenes), facilitating model evaluation and deployment.

\subsection{Multimodal Feature Input and Data Transfer}

The preprocessed visual features \( F_v \) and text features \( F_t \) are concatenated to form a unified input \( F = [F_v; F_t] \in \mathbb{R}^{d_v + d_t} \), which is transmitted to the Qwen2-VL model for task-specific processing. To optimize data transfer efficiency, the framework employs batch processing with a batch size of 32 during training, leveraging GPU parallelization to reduce latency. The input pipeline is further optimized using asynchronous data loading and prefetching, minimizing I/O bottlenecks. The concatenated features are passed through a projection layer to align their dimensions, ensuring seamless integration into the Qwen2-VL transformer architecture. This step enhances scalability for large datasets and supports real-time inference, critical for autonomous driving applications.

\subsection{Multimodal Feature Fusion with LoRA}
Traditional multimodal models require full-parameter fine-tuning to adapt to tasks like 3D open-set perception, incurring significant computational costs and limiting deployment on resource-constrained devices. To address this, We adopt a feature fusion approach using low-rank adaptation (LoRA) integrated with Qwen2-VL, enabling efficient task adaptation with minimal parameter updates.

The LoRA module modifies the pretrained weight matrix \( W \in \mathbb{R}^{d \times d} \) of the Qwen2-VL transformer layers by introducing a low-rank update \( \Delta W = BA \), where \( A \in \mathbb{R}^{r \times d} \), \( B \in \mathbb{R}^{d \times r} \), and \( r \ll d \) is the rank (set to 16 in our implementation). The parameter count is reduced from \( d^2 \) to \( 2dr \), significantly lowering computational overhead. A scaling factor \( \alpha = 32 \) adjusts the magnitude of the update, and the modified weights are computed as:
\begin{equation}
W' = W + \alpha \cdot \Delta W
\end{equation}

The input features \( F = [F_v; F_t] \) are processed through the Qwen2-VL transformer, which consists of alternating self-attention and feed-forward layers. The transformer outputs a fused feature representation, which is fed into a multilayer perceptron (MLP) with three hidden layers (512, 256, and 128 units, respectively, with ReLU activations) to generate the 3D bounding box parameters \( [x, y, z, l, w, h, \theta] \). The MLP is trained to map the fused features to the 7-dimensional output space, ensuring accurate prediction of pedestrian locations, sizes, and orientations.

To further optimize performance, LoRA is applied only to the self-attention layers of Qwen2-VL, reducing the number of trainable parameters to approximately 0.1\% of the original model. This lightweight adaptation enables deployment on embedded systems with limited computational resources (e.g., NVIDIA Jetson platforms). The feature fusion process is enhanced by a cross-modal attention mechanism within Qwen2-VL, which dynamically weights visual and text features based on their relevance to the task, improving semantic alignment and robustness to diverse inputs.

\subsection{Two-Stage Loss Optimization}
Generating high-precision 3D bounding boxes requires balancing semantic alignment between multimodal inputs and geometric accuracy of the output. A single loss function struggles to optimize both objectives simultaneously, particularly in complex traffic scenarios with diverse objects and occlusion. We propose a two-stage loss optimization strategy that combines Mean Squared Error (MSE) loss for preliminary feature alignment and Intersection over Union (IoU) loss for refined bounding box alignment, ensuring robust performance across open-set scenarios.

\subsection{Stage 1: Preliminary Feature Alignment}

In the first stage, the framework focuses on aligning the fused multimodal features with the semantic content of the ground-truth bounding boxes. The Qwen2-VL model, enhanced with LoRA, generates initial 3D bounding box predictions \( P_i = [x_i, y_i, z_i, l_i, w_i, h_i, \theta_i] \) for each sample \( i \). These predictions are embedded into a semantic feature space using a projection head (a linear layer mapping to \( \mathbb{R}^{128} \)), producing semantic features \( f_{\text{pred}}^i \). The ground-truth bounding boxes \( G_i = [x_i^*, y_i^*, z_i^*, l_i^*, w_i^*, h_i^*, \theta_i^*] \) are similarly embedded to produce \( f_{\text{gt}}^i \). The MSE loss quantifies the semantic discrepancy between predicted and ground-truth features:
\begin{equation}
\mathcal{L}_{\text{MSE}} = \frac{1}{N} \sum_{i=1}^N \left\| f_{\text{pred}}^i - f_{\text{gt}}^i \right\|^2
\end{equation}

where \( N \) is the number of samples in a batch. The MSE loss drives rapid convergence by optimizing the model to capture high-level semantic relationships between the input prompt and image, ensuring that the initial bounding box predictions are semantically consistent with the task requirements. This stage typically continues for 50 epochs, with a learning rate of \( 10^{-4} \) using the AdamW optimizer, until the semantic alignment stabilizes (indicated by a plateau in validation MSE loss).

\subsection{Stage 2: Refined Bounding Box Alignment}
In the second stage, the framework refines the geometric accuracy of the 3D bounding boxes using IoU loss, which directly optimizes the overlap between predicted and ground-truth boxes. For each sample \( i \), the IoU is computed as:

\begin{equation}
{IoU}_i = \frac{P_i \cap {G_i}}{P_i \cup {G_i}}
\end{equation}

where \( P_i \) and \( G_i \) represent the 3D volumes of the predicted and ground-truth bounding boxes, respectively, calculated based on their center, size, and yaw parameters. The IoU loss is defined as:

\begin{equation}
\mathcal{L}_{\text{IoU}} = 1 - \text{IoU}_i
\end{equation}

This loss is aggregated across all samples in a batch:

\begin{equation}
\mathcal{L}_{\text{IoU}} = \frac{1}{N} \sum_{i=1}^N (1 - \text{IoU}_i)
\end{equation}

The IoU loss optimizes the precise localization of the bounding box center (\( x, y, z \)), size (\( l, w, h \)), and yaw angle (\( \theta \)), ensuring high geometric accuracy in complex scenarios with occlusions or varying object scales. This stage runs for an additional 50 epochs, with a reduced learning rate of \( 10^{-5} \) to fine-tune the model parameters.

The overall loss combines both stages with dynamic weighting:

\begin{equation}
\mathcal{L} = \lambda_1 \mathcal{L}_{\text{MSE}} + \lambda_2 \mathcal{L}_{\text{IoU}}
\end{equation}

where the coefficients \( \lambda_1 \) and \( \lambda_2 \) are adjusted based on the training stage:
- Stage 1 (epochs 1–50): \( \lambda_1 = 1.0 \), \( \lambda_2 = 0.0 \), focuing solely on MSE loss for semantic alignment.
- Stage 2 (epochs 51–100): \( \lambda_1 = 0.2 \), \( \lambda_2 = 0.8 \), emphasizing IoU loss for geometric refinement while retaining some influence of MSE loss to maintain semantic consistency.

The transition threshold is set at 100 epochs, determined empirically based on validation performance. To evaluate the effectiveness of the two-stage optimization, we adopt the mean Intersection over Union (mIoU), defined as the mean IoU across all samples:

\begin{equation}
\text{mIoU} = \frac{1}{N} \sum_{i=1}^N \text{IoU}_i
\end{equation}

The mIou metric quantifies the overall geometric accuracy of the predicted bounding boxes, providing a standardized measure for model evaluation. Qualitative analysis is performed by visualizing predicted versus ground-truth bounding boxes in 3D space, using tools like Open3D or PyVista, to assess semantic alignment (Stage 1) and geometric precision (Stage 2).

\subsection{Open-Set Strategy}
The proposed framework incorporates explicit strategies for handling unseen categories, thereby enabling robust open-set perception—a critical capability for autonomous driving in dynamic, open-world environments. Traditional closed-set models often assume a fixed set of known object classes during both training and inference, which limits their applicability in real-world scenarios where unexpected objects frequently appear. In contrast, this framework is designed to recognize and localize novel or previously unseen objects, such as strollers, animals, or construction equipment, by leveraging generalized representations and uncertainty-aware mechanisms. This open-set design significantly enhances the model’s flexibility and resilience, allowing it to function safely and reliably in ever-changing traffic conditions.

During the evaluation phase, we assess the framework’s generalization ability using datasets that deliberately include unseen object categories not present during training. The model's effectiveness is measured by its ability to produce accurate 3D bounding boxes for these novel objects, with performance quantitatively evaluated using the mean Intersection over Union (mIoU) metric. As shown in Table 3, the framework maintains strong localization performance even for unseen categories, demonstrating its capacity to extend beyond learned object classes. This open-set recognition capability is essential for real-world autonomous driving systems, where encountering diverse and unpredictable objects is the norm rather than the exception.

\section{Experiment}

\subsection{Experiment Setup}
In this paper, we evaluate the proposed model VLM-3D using the nuScenes dataset. We utilized the complete nuScenes dataset, comprising 1000 driving scenes collected from Boston and Singapore, covering diverse urban environments with complex traffic, pedestrian interactions, and varying weather conditions. Its multimodal features, including camera images, LiDAR point clouds, and radar scans, along with high-quality 3D bounding box annotations, support tasks such as object detection, tracking, and semantic segmentation.
The data processing pipeline extracts LiDAR point clouds and images from six cameras (front, front-right, front-left, back, back-left, back-right). 3D bounding boxes are transformed from the global coordinate system to the LiDAR coordinate system using transformation matrices derived from ego pose and sensor calibration data, computing their center, size, and yaw angle. Bounding box corners are projected onto the image plane, retaining only those with positive Z-coordinates and within the image boundaries, ensuring data accuracy.
\subsection{Experiment Environment}

The experiments were conducted on a high-performance computing server with the following configuration: an AMD EPYC 7642 processor featuring 48 cores and 96 threads, operating at 3.293 GHz; an NVIDIA A100-PCIE-40GB GPU; and 512 GB of DDR4 ECC memory (16 × 32 GB, 2666 MHz). The system was equipped with CUDA version 12.1 and ran on CentOS Linux 7 (Core) with kernel version 3.10.0.

\begin{table}[!htb]
\centering
\renewcommand{\arraystretch}{1.1} 
\setlength{\tabcolsep}{10pt} 
\begin{tabular}{m{1cm}rrrl} 
\toprule
Stage & Epoch & Accuracy & Recall & F1 \\
\midrule
\multirow{21}{1.2cm}[0pt]{Stage1} 
      & 1  & 30.2\% & 31.4\% & 30.8\% \\
      & 5  & 32.0\% & 45.2\% & 37.5\% \\
      & 7  & 29.2\% & 53.2\% & 37.7\% \\
      & 9  & 53.9\% & 45.6\% & 49.4\% \\
      & 11 & 59.9\% & 40.8\% & 48.5\% \\
      & 13 & 59.5\% & 43.4\% & 50.2\% \\
      & 15 & 59.8\% & 45.5\% & 51.7\% \\
      & 17 & 47.7\% & 45.5\% & 46.6\% \\
      & 19 & 68.4\% & 43.9\% & 53.5\% \\
      & 21 & 59.3\% & 41.9\% & 49.1\% \\
      & 23 & 70.5\% & 41.2\% & 52.0\% \\
      & 25 & 56.0\% & 45.5\% & 50.2\% \\
      & 27 & 56.7\% & 46.6\% & 51.1\% \\
      & 29 & 56.0\% & 45.5\% & 50.2\% \\
      & 31 & 66.0\% & 41.1\% & 50.6\% \\
      & 33 & 67.4\% & 42.8\% & 52.3\% \\
\midrule
\multirow{9}{1.8cm}[0pt]{Stage 2} 
      & 35 & 64.4\% & 44.2\% & 52.4\% \\
      & 37 & 67.1\% & 44.5\% & 53.5\% \\
      & 39 & 66.5\% & 45.1\% & 53.7\% \\
      & 41 & 63.4\% & 45.2\% & 52.8\% \\
      & 43 & 64.0\% & 44.4\% & 52.5\% \\
      & 45 & 63.4\% & 45.3\% & 52.8\% \\
      & 47 & 66.4\% & 44.1\% & 53.0\% \\
      & 49 & 70.3\% & 40.2\% & 51.2\% \\
      & 51 & 68.8\% & 40.7\% & 51.1\% \\
\bottomrule
\end{tabular}
\caption{Accuracy, Recall, and F1 scores for different epochs in Stage 1 and Stage 2.}
\label{table:stage_metrics}
\end{table}
\subsection{Evaluation Metrics}
To evaluate the proposed framework, we assess its performance using five key metrics: Accuracy, Recall, F1 Score,and mean Intersection over Union (mIoU). These metrics provide a comprehensive evaluation of the model's classification, detection, and open-set robustness.

Accuracy measures the proportion of correctly predicted samples:

\begin{equation}
\text{Accuracy} = \frac{\text{TP} + \text{TN}}{\text{TP} + \text{TN} + \text{FP} + \text{FN}}
\end{equation}

where TP, TN, FP, and FN represent true positives, true negatives, false positives, and false negatives, respectively. Accuracy, defined as the ratio of correctly predicted samples to the total number of samples, is a commonly used evaluation metric due to its intuitive interpretation and ease of computation. Therefore, complementary metrics such as precision, recall, and F1-score are often employed alongside accuracy to provide a more comprehensive evaluation of model performance, especially in imbalanced classification tasks.

Recall, or true positive rate, evaluates the model's ability to detect positive samples:

\begin{equation}
\text{Recall} = \frac{\text{TP}}{\text{TP} + \text{FN}}
\end{equation}

High recall is critical for applications where failing to identify positive instances can lead to severe consequences. For example, in pedestrian detection within autonomous driving systems, missing a pedestrian (i.e., a false negative) could result in safety hazards or even life-threatening situations. In such safety-critical scenarios, recall reflects the model’s ability to capture all actual positive cases, and a low recall may indicate that the system overlooks important detections. Therefore, models deployed in these contexts are typically optimized to achieve high recall, even at the cost of increased false positives, which are often more tolerable than false negatives. This trade-off emphasizes the importance of recall over other metrics in applications such as medical diagnosis, intrusion detection, and surveillance systems, where the cost of omission far outweighs the cost of false alarms.

The F1 Score serves as a harmonic mean of precision and recall, providing a single metric that balances both aspects of classification performance. Unlike the arithmetic mean, the harmonic mean penalizes extreme values, ensuring that a high F1 Score can only be achieved when both precision and recall are comparably high. This makes the F1 Score particularly useful in imbalanced classification tasks, where a model might exhibit high precision but low recall, or vice versa.It is widely adopted in domains such as information retrieval, natural language processing, and biomedical signal classification, where the balance between correctly detecting relevant instances and avoiding false alarms is crucial:

\begin{equation}
\text{F1 Score} = 2 \cdot \frac{\text{Precision} \cdot \text{Recall}}{\text{Precision} + \text{Recall}}
\end{equation}

It is suitable for imbalanced datasets, providing a robust measure of overall performance.

The mean Intersection over Union (mIoU) is a widely used metric for evaluating semantic segmentation and other tasks requiring precise spatial localization of objects. Unlike pixel-level accuracy, which can be biased toward dominant classes, mIoU provides a more balanced evaluation by treating each category equally, making it especially suitable for datasets with class imbalance. Moreover, mIoU captures both classification correctness and spatial alignment, offering a robust measure of how well the model delineates object boundaries and distinguishes adjacent or overlapping regions in complex scenes. As such, it has become a standard benchmark in computer vision tasks like autonomous driving, remote sensing, and medical image segmentation:

\begin{equation}
\text{mIoU} = \frac{1}{C} \sum_{c=1}^C \text{IoU}_c
\end{equation}

where \( C \) is the number of categories. Higher mIoU indicates better spatial precision, critical for 3D bounding box generation.

\begin{figure*}[htbp]
    \centering
    \parbox{\textwidth}{\centering{Prompt: Please detect the * in the image and output 3D box center (x, y, z) , size (length, width, height) , yaw in radians.}}
    \vspace{0.5em}
    \subfigure[Construction]{
        \includegraphics[width=2in]{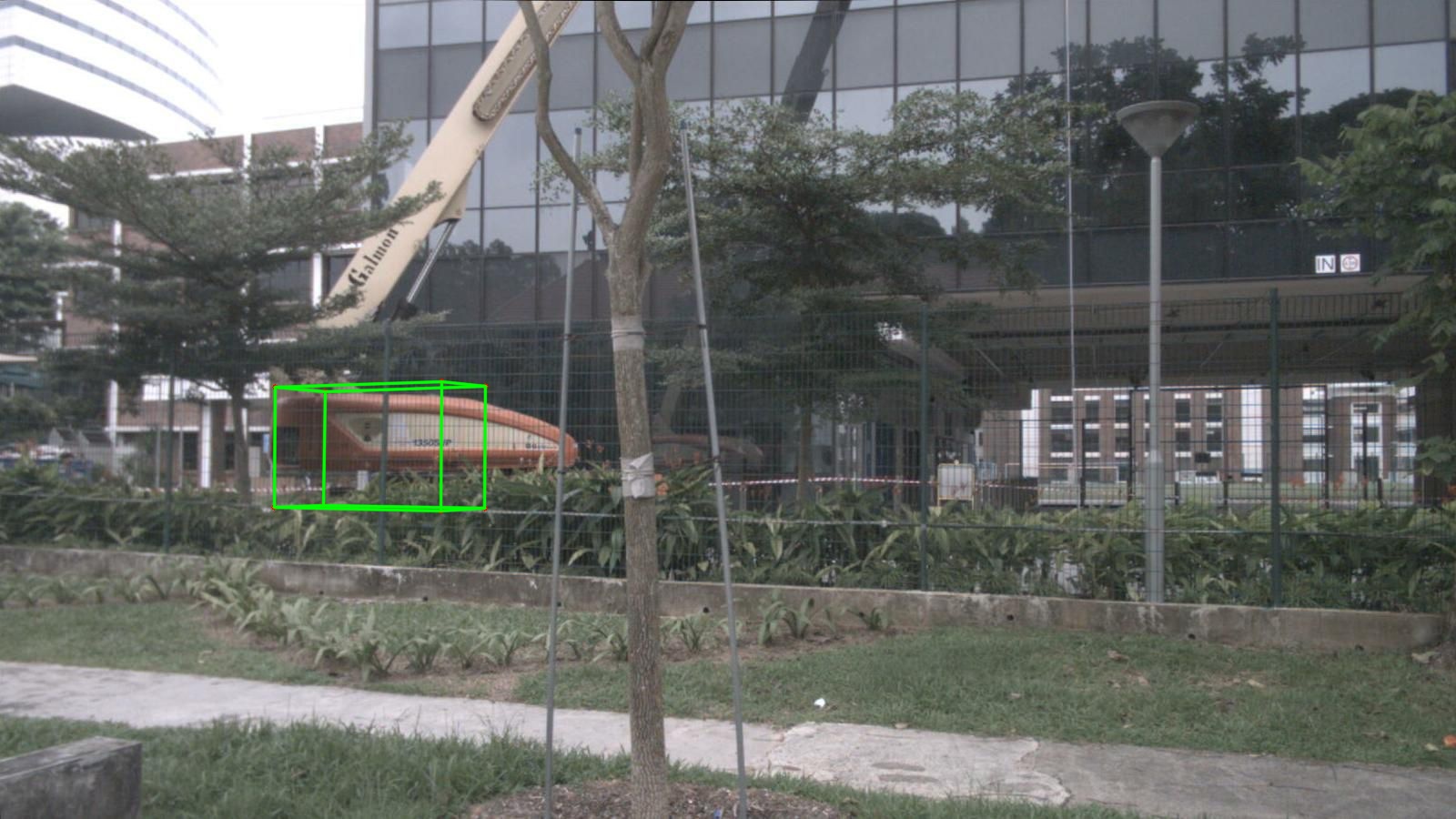}
        \label{fig:construction}
    }
    \hspace{0.5em}
    \subfigure[Rigid]{
        \includegraphics[width=2in]{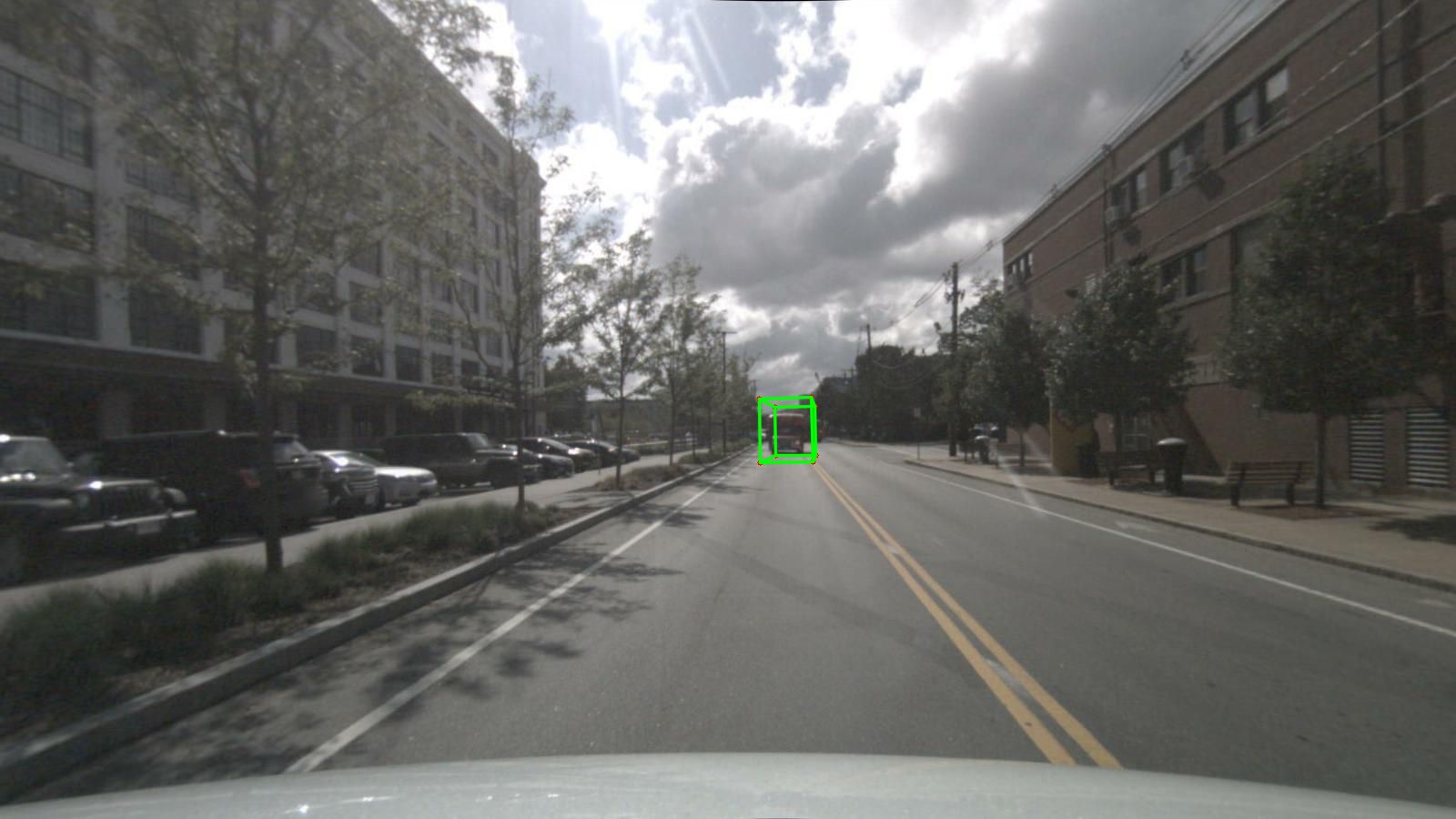}
        \label{fig:rigid1}
    }
    \hspace{0.5em}
    \subfigure[Adult]{
        \includegraphics[width=2in]{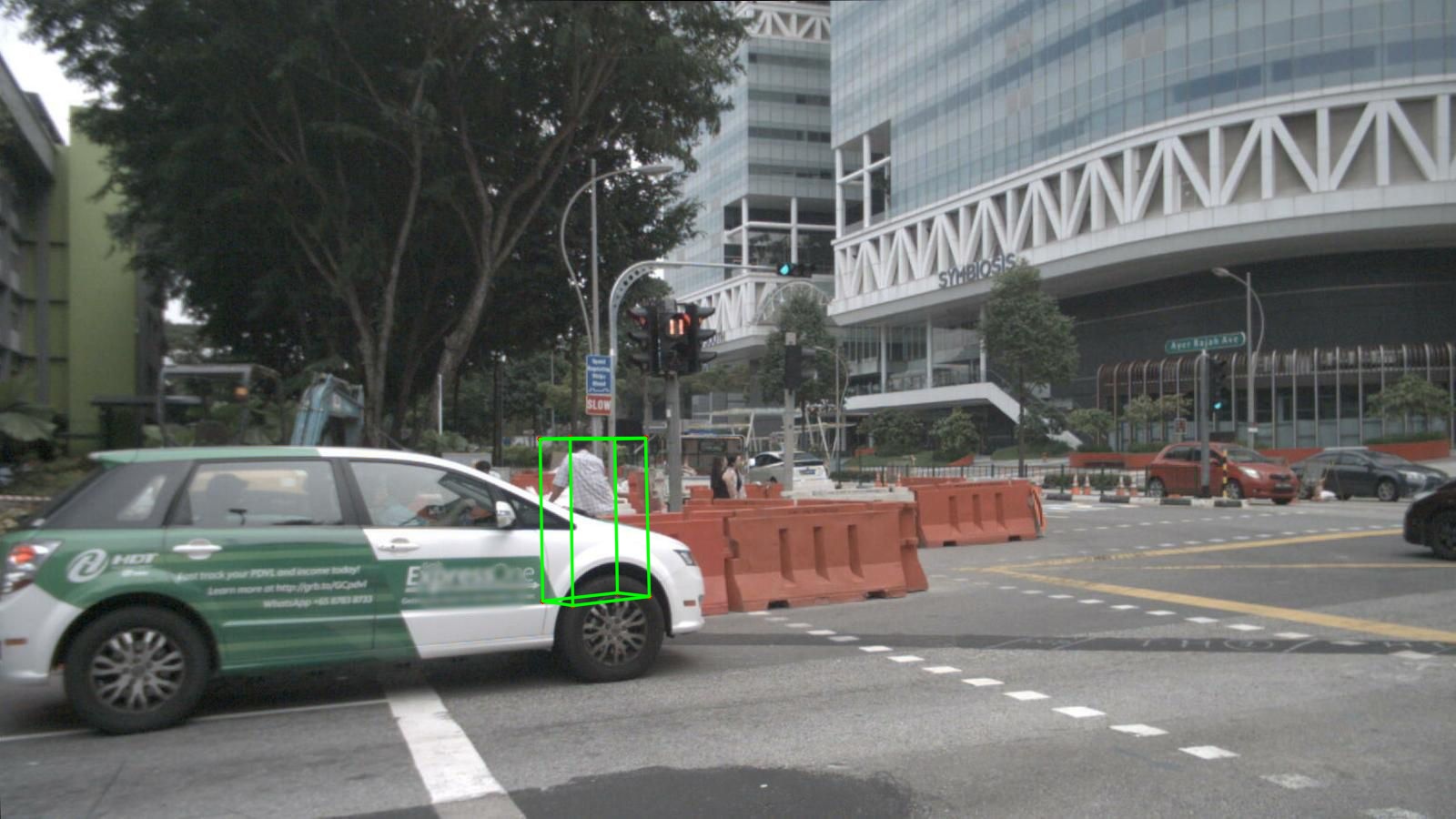}
        \label{fig:adult}
    }
    \quad
    \subfigure[Rigid]{
        \includegraphics[width=2in]{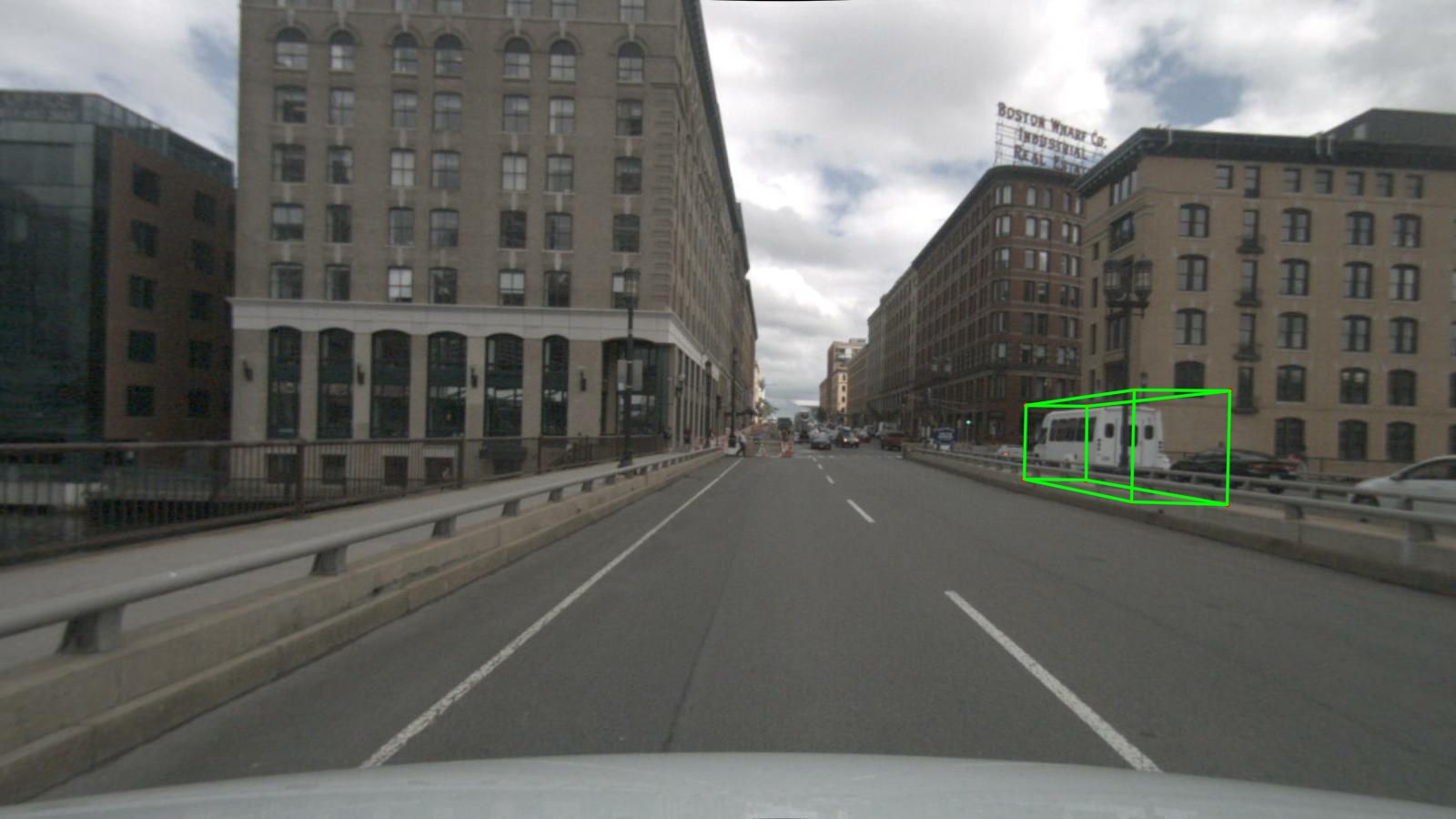}
        \label{fig:rigid2}
    }
    \hspace{0.5em}
    \subfigure[Construction worker]{
        \includegraphics[width=2in]{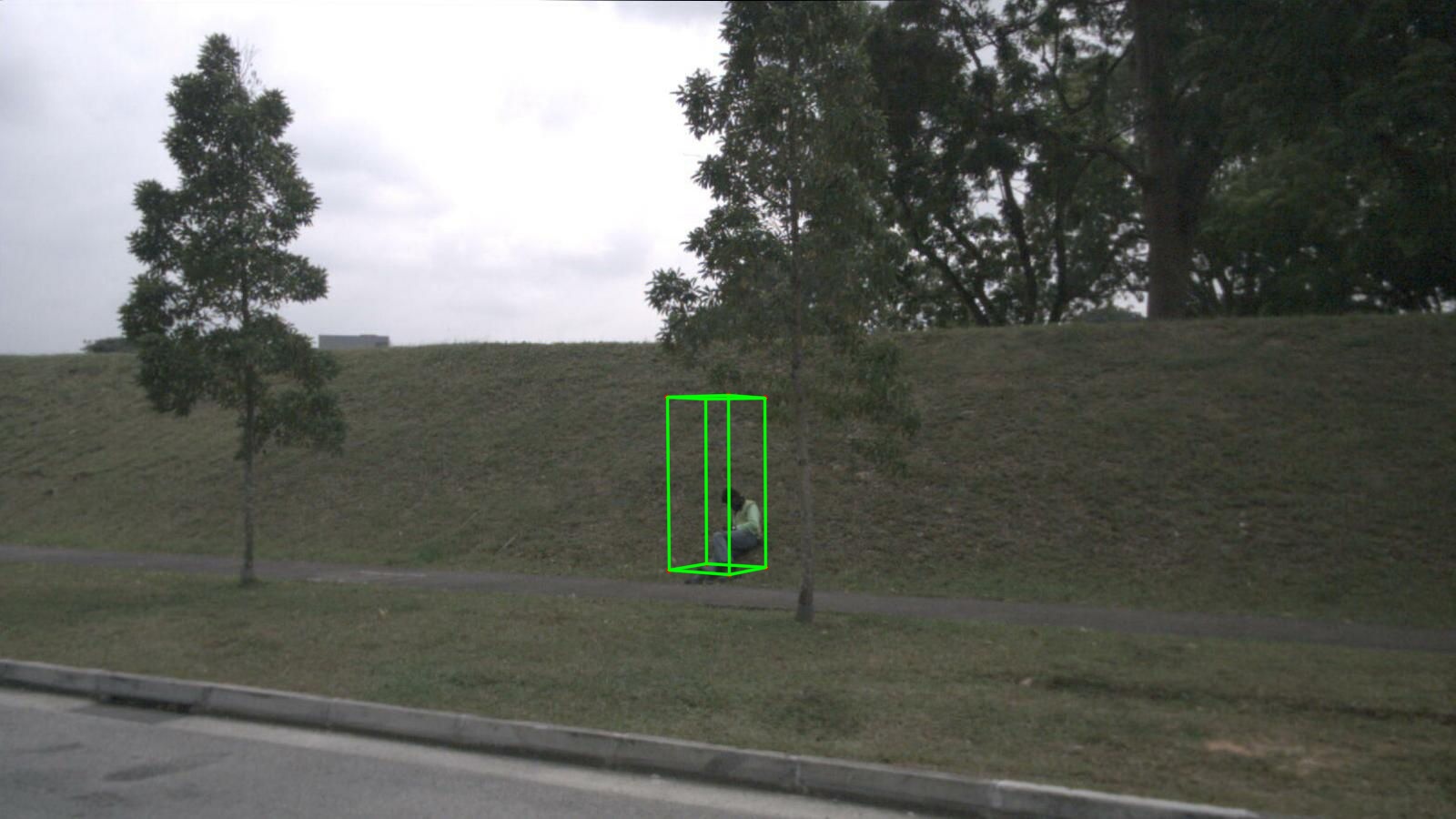}
        \label{fig:construction_worker}
    }
    \hspace{0.5em}
    \subfigure[Animal]{
        \includegraphics[width=2in]{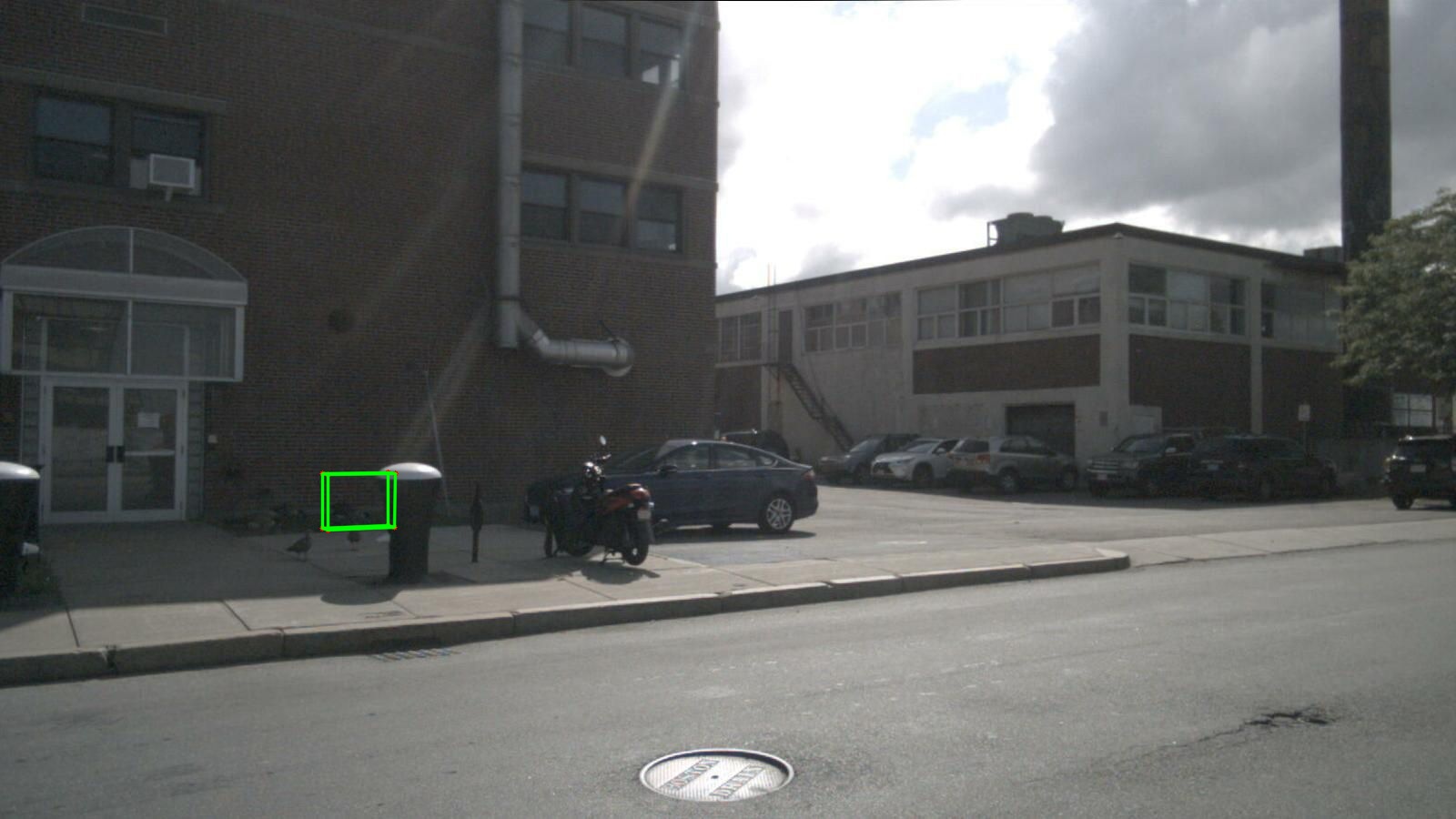}
        \label{fig:animal}
    }
    \caption{Visualization of object detection results for open-set categories.}
    \label{fig:open_set_detection}
\end{figure*}

\subsection{Result}

\begin{table}[!htb]
\centering
\small 
\renewcommand{\arraystretch}{1.1} 
\begin{tabular}{lrlr}
\toprule
\multicolumn{2}{c}{Training Set} & \multicolumn{2}{c}{Test Set} \\
\cmidrule(lr){1-2} \cmidrule(lr){3-4}
Category & IoU & Category & IoU \\
\midrule
car & 0.2354 & car & 0.2275 \\
adult & 0.2469 & adult & 0.2521 \\
trafficcone & 0.2412 & trafficcone & 0.2109 \\
truck & 0.2180 & truck & 0.1968 \\
barrier & 0.2429 & barrier & 0.2353 \\
construction & 0.2145 & construction & 0.1721 \\
pushable\_pullable & 0.1839 & pushable\_pullable & 0.2803 \\
rigid & 0.2383 & rigid & 0.2086 \\
construction\_worker & 0.2869 & construction\_worker & 0.2674 \\
motorcycle & 0.2469 & motorcycle & 0.2187 \\
bicycle & 0.2224 & bicycle & 0.2189 \\
trailer & 0.2177 & trailer & 0.1384 \\
bicycle\_rack & 0.2002 & bicycle\_rack & 0.2270 \\
child & 0.2071 & child & 0.2822 \\
bendy & 0.2485 & bendy & 0.2451 \\
wheelchair & 0.2997 & wheelchair & 0.2301 \\
mIoU & 0.2344 & mIoU & 0.2257 \\
\bottomrule
\end{tabular}
\caption{IoU for different categories in training and test sets.}
\label{table:iou}
\end{table}

We evaluated the proposed end-to-end 3D open-set perception framework on the NuScenes dataset, demonstrating that the joint semantic and 3D geometric loss design improves prediction accuracy by 12.8\%, validating the effectiveness of our approach. Table 1 summarizes the Accuracy and Recall metrics for Stage 1 and Stage 2 across multiple training epochs. In Stage 1, the model is trained with a joint semantic loss, achieving a steady increase in Accuracy from 30.2\% at Epoch 1 to a peak of 70.5\% at Epoch 23. However, after reaching 67.4\% at Epoch 33, the loss begins to increase, potentially due to overfitting or variations in the training data distribution.

To address the performance fluctuations observed in Stage 1 after Epoch 33, we introduced a joint optimization strategy combining semantic loss and 3D geometric IoU loss, initiating Stage 2 training from Epoch 33. As shown in Table 1, Stage 2 consistently outperforms Stage 1, with Accuracy improving from 64.4\% at Epoch 35 to a peak of 70.3\% at Epoch 49, stabilizing at 68.8\% by Epoch 51. Recall ranges between 40.2\% and 45.3\%, slightly higher than Stage 1 in later epochs, indicating that the joint loss optimization effectively reduces false negatives. The incorporation of geometric feature loss in Stage 2 significantly enhances bounding box predictions in open-set scenarios, underscoring the robustness of the proposed framework.

Table 2 presents the Intersection over Union and instance counts for various categories in the training and test sets of the NuScenes dataset. The mean IoU is 0.2366 for the training set and 0.2352 for the test set, indicating consistent performance across both sets. Categories such as wheelchair and construction\_worker achieve high IoU in the training set, while personal\_mobility  and debris perform strongly in the test set, reflecting robust detection for sparse categories. However, categories like police  and trailer show lower IoU, suggesting challenges in detecting underrepresented or geometrically complex objects. These results highlight the framework’s capability to handle diverse object categories in open-set 3D detection.

\subsubsection{Visualization Analysis}
Figure 2 presents a 2×3 grid visualizing object detection results, showing 3D bounding boxes for construction equipment Figure~\ref{fig:construction}, rigid objects Figure~\ref{fig:rigid1}, Figure~\ref{fig:rigid2}, and a construction worker Figure~\ref{fig:construction_worker}. These subfigures validate the model’s generalization to unseen categories in dynamic autonomous driving scenarios, complementing the IoU metrics in Table~\ref{table:iou}, and demonstrate robust performance on diverse objects.

\section{Conclusion}
This study proposes an end-to-end 3D open-set perception framework that significantly enhances detection performance on the NuScenes dataset through a joint semantic and 3D geometric loss design. Table 1 demonstrates that Stage 1 achieves substantial accuracy improvements using joint semantic loss, while Stage 2, incorporating a joint optimization of semantic and 3D geometric IoU loss, further enhances model robustness and prediction stability, particularly excelling in bounding box prediction for open-set scenarios. Table 2 indicates that the framework exhibits strong detection capabilities across diverse categories, with notable robustness for sparse categories, validating the effectiveness of low-rank adaptation and dynamic feature fusion in 3D object detection and providing a reliable solution for open-set detection tasks.

\section{Acknowledgments}

This work was supported by the National Key R\&D Program of China,Project “Development of Large Model Technology and Scenario Library Construction for Autonomous Driving Data Closed-Loop" (Grant No. 2024YFB2505501), and the Guangxi Key Scientific and Technological Project, Project “Research and Industrialization of High-Performance and Cost-Effective Urban Pilot Driving Technologies”(Grant No.Guangxi Science and Technology AA24206054).

\end{document}